\documentclass[letterpaper, 10 pt, conference]{ieeeconf}  

\IEEEoverridecommandlockouts                              

\usepackage{mymacros}
\usepackage[para]{threeparttable}
\usepackage{caption}
\usepackage[font=small]{caption}

\usepackage{subcaption}
\usepackage{multirow}

\let\labelindent\relax
\usepackage{enumitem}

\title{\LARGE \bf FastDepth: Fast Monocular Depth Estimation on Embedded Systems}

\author{
  \thanks{Authors are with the Massachusetts Institute of Technology, Cambridge, MA 02139, USA.
  Emails: \texttt{\{dwofk, fcma, tjy, sertac, sze\}@mit.edu.}
  \textbf{Project website: \url{http://fastdepth.mit.edu}}}
  Diana Wofk$^*$, Fangchang Ma$^*$, Tien-Ju Yang, Sertac Karaman, Vivienne Sze
}

\begin{document}
\bstctlcite{IEEEexample:BSTcontrol} 
\maketitle
\thispagestyle{empty}
\pagestyle{empty}


\begin{abstract}

Depth sensing is a critical function for robotic tasks such as localization, mapping and obstacle detection. There has been a significant and growing interest in depth estimation from a single RGB image, due to the relatively low cost and size of monocular cameras. 
However, state-of-the-art single-view depth estimation algorithms are based on fairly complex deep neural networks that are too slow for real-time inference on an embedded platform, for instance, mounted on a micro aerial vehicle.  
In this paper, we address the problem of fast depth estimation on embedded systems. We propose an efficient and lightweight encoder-decoder network architecture and apply network pruning to further reduce computational complexity and latency. In particular, we focus on the design of a low-latency decoder. Our methodology demonstrates that it is possible to achieve similar accuracy as prior work on depth estimation, but at inference speeds that are an order of magnitude faster. Our proposed network, FastDepth, runs at 178 fps on an NVIDIA Jetson TX2 GPU and at 27 fps when using only the TX2 CPU, with active power consumption under 10 W. FastDepth achieves close to state-of-the-art accuracy on the NYU Depth v2 dataset.
To the best of the authors' knowledge, this paper demonstrates real-time monocular depth estimation using a deep neural network with the lowest latency and highest throughput on an embedded platform that can be carried by a micro aerial vehicle.

\end{abstract}

\section{Introduction}
\label{sec:introduction}

Depth sensing is essential to many robotic tasks, including mapping, localization, and obstacle avoidance. Existing depth sensors (e.g., LiDARs, structured-light sensors, etc.) are typically bulky, heavy, and have high power consumption. These limitations make them unsuitable for small robotic platforms (e.g., micro aerial and mini ground vehicles), which motivates depth estimation using a monocular camera, due to its low cost, compact size, and high energy efficiency. 

Past research on monocular depth estimation has focused almost exclusively on improving accuracy, resulting in computation-intensive algorithms that cannot be readily adopted in robotic systems. Since most systems are not only limited in compute resources, but are also subject to latency constraints, a key challenge is balancing the computation and runtime cost with the accuracy of the algorithm.

Current state-of-the-art depth estimation algorithms rely on deep learning based methods, and while these achieve significant improvement in accuracy, they do so at the cost of increased computational complexity. Prior research on designing fast and efficient networks has primarily focused on encoder networks for tasks such as image classification and object detection~\cite{sze2017efficient}. In these applications, the input is an image (pixel-based), and the output is reduced to a label (an object class and position). To the best of our knowledge, little effort has been put into the efficient design of \emph{both} encoder and decoder networks (i.e., auto-encoder networks) for tasks such as depth estimation, where the output is a dense image. In particular, reducing decoder complexity poses an additional challenge since there is less information reduction at each of the decoding layers and the decoder's output is high dimensional.

\begin{figure}[t]
\centering
\includegraphics[width=0.95\linewidth]{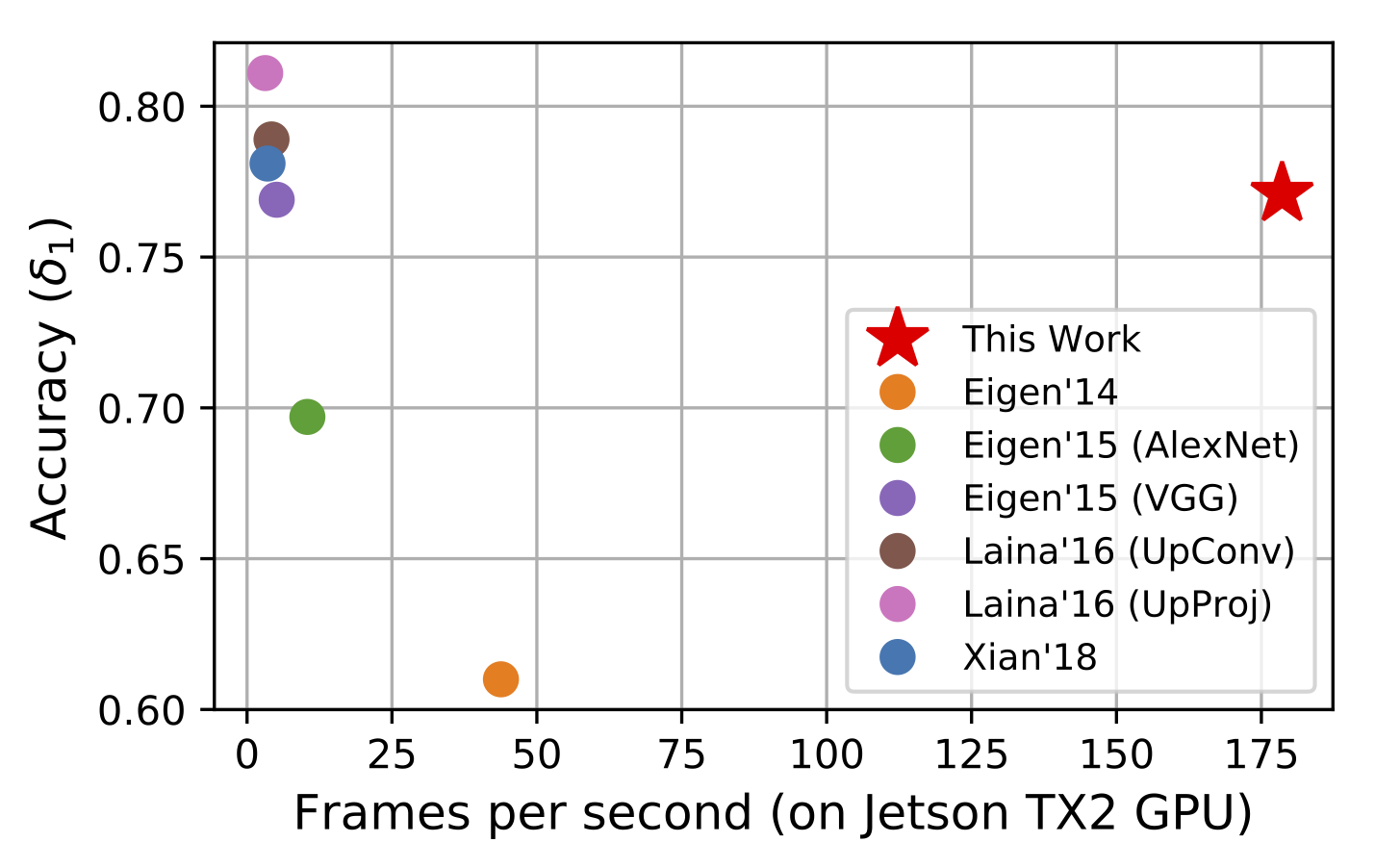}
\caption{Accuracy vs. runtime (in fps) on an NVIDIA Jetson TX2 GPU for various depth estimation algorithms. Top right represents the desired characteristics of a depth estimation network design: high throughput and high accuracy.}
\label{fig:frontier}
\end{figure}

To address these challenges, this paper presents a low latency, high-throughput, high-accuracy depth estimation algorithm running on embedded systems. We propose an efficient encoder-decoder network architecture with a focus on low latency design. Our approach employs \mobilenet~\cite{howard2017mobilenets} as an encoder and nearest neighbor interpolation with depthwise separable convolution in the decoder. We apply state-of-the-art network pruning, NetAdapt~\cite{yang2018netadapt}, and use the TVM compiler stack~\cite{chen2018tvm} to further reduce inference runtime on a target embedded platform. We show that our low latency network design, FastDepth, can perform real-time depth estimation on the NVIDIA Jetson TX2~\cite{jetsontx2}, operating at over 120 frames per second (fps) on the TX2 GPU (see~\prettyref{fig:frontier}) and at over 25 fps on the TX2 CPU only\footnote{This throughput is achieved with a batch size of one and 32-bit floating point precision. Throughput can be increased by using a larger batch size (at the cost of higher latency), and/or reducing bitwidths through quantization.}, with active power consumption under 10 W. The attained throughput is an order of magnitude higher than prior work on depth estimation, with only a slight loss of accuracy; FastDepth achieves a $\delta_1$ accuracy\footnote{Accuracy metrics are defined in \prettyref{sec:exp-setup}.} of 77.1\% on NYU Depth v2.

The low latency and high throughput attainable with our network is motivated by practical robotic systems, where multiple programs (such as localization, mapping, motion planning, control, and potentially other perception tasks) all run in parallel. Each program demands a certain amount of computational resources, and consequently, the CPU/GPU is not dedicated to the depth estimation task alone. A lower latency network design enables real-time performance even with a limited computing budget. 

In summary, this paper demonstrates real-time monocular depth estimation using a deep neural network that achieves the lowest latency and highest throughput on an embedded platform that can be carried by a micro aerial vehicle.

\section{Related Work} 
\label{sec:related_work}

In this section, we summarize past research done on depth estimation, efficient neural networks, and network pruning.

\subsection{Monocular Depth Estimation}

Depth estimation from a single color image has been an active research topic in both the robotics and computer vision communities for over a decade. Early works on depth estimation using RGB images captured by a monocular camera usually relied on hand-crafted features and probabilistic graphical models. For instance, \citet{saxena2006learning} estimated the absolute scales of different image patches and inferred depth using a Markov Random Field model. Non-parametric approaches~\cite{karsch2012depth,konrad20122d,karsch2014depthtransfer,liu2014discrete} were also exploited to estimate the depth of a query image by combining the depths of images with similar photometric content retrieved from a database. Since then, depth estimation has evolved from using simple handcrafted feature representations~\cite{saxena2006learning} to modern deep learning based approaches~\cite{eigen2014depth,laina2016deeper,ummenhofer2016demon,fu2018deep}. 

State-of-the-art RGB-based depth estimation methods use deep learning based methods to train a convolution neural network using large-scale datasets~\cite{liu2015deep,eigen2015predicting,laina2016deeper}. \citet{eigen2014depth} suggested a two-stack convolutional neural network (CNN), with one stack predicting the global coarse scale and the other stack refining local details. \citet{eigen2015predicting} further incorporated other auxiliary prediction tasks into the same architecture. \citet{liu2015deep} combined a deep CNN and a continuous conditional random field, and attained visually sharper transitions and local details.
\citet{laina2016deeper} developed a deep residual network based on ResNet~\cite{he2016deep} and achieved higher accuracy than \cite{liu2015deep,eigen2015predicting}. \citet{qi2018geonet} trained networks to estimate both the depth and the normals, as a way to address the problem of blurriness in predictions. Semi-supervised~\cite{kuznietsov2017semi} and unsupervised learning~\cite{zhou2017unsupervised,garg2016unsupervised,godard2016unsupervised} setups have also been explored for disparity image prediction. For instance, \citet{godard2016unsupervised} formulated disparity estimation as an image reconstruction problem, where neural networks were trained to warp left images to match the right. \citet{mancini2016fast} proposed a CNN that took both RGB images and optical flow images as input to predict distance. Fusion of RGB images and sparse depth measurements~\cite{ma2017sparse, ma2018self} at early stages also improved the accuracy of depth estimation.

All of these methods focus heavily on attaining higher accuracy at increased complexity and runtime cost, with diminishing accuracy improvement. For instance, the $\delta_1$ accuracy of depth estimation on the NYU Depth V2 dataset~\cite{silberman2012indoor} saturates at around 82\% in recent years~\cite{laina2016deeper,qi2018geonet}.

\subsection{Efficient Neural Networks}

There has been significant effort in prior work to design efficient neural networks. As an example, for image classification, \mobilenet~\cite{howard2017mobilenets} achieves similar accuracy as \vgg{-16}~\cite{simonyan2014very} but has $2.7$ times fewer multiply-and-accumulate operations (\mac) and $32.9$ times fewer weights. For object detection, SSD~\cite{liu2016ssd} is $6.6$ times faster than Faster-RCNN~\cite{ren2015fasterrcnn} with a higher mean average precision (mAP).

However, most previous work in this space has focused on encoder networks that reduce an input image into a label. Designing efficient neural networks for applications requiring pixel-based results, where an encoder network is followed by a decoder network, has been less explored. As will be shown in~\prettyref{fig:progression}, in existing designs that achieve close to state-of-the-art accuracy on the depth estimation task, the decoder largely dominates inference runtime. In our work, we emphasize efficient encoder-decoder network design. In particular, our usage of depthwise separable convolution in the decoder differentiates us from existing approaches and enables us to develop an architecture in which the decoder no longer dominates inference runtime.

\subsection{Network Pruning}

Hand-crafted networks are usually over-parameterized due to the ease of training, which reduces network efficiency. To address this problem, network pruning, such as~\cite{leoptimal,gordon2018morphnet,cvpr2017-yang-energy-aware-pruning,yang2018netadapt}, is widely used to identify and remove redundant parameters and computation. However, these pruning methods are mainly applied on encoder networks. In this work, we adopt a state-of-the-art algorithm, NetAdapt~\cite{yang2018netadapt}, to demonstrate how pruning can improve the efficiency of \emph{both} encoder and decoder networks used in a depth estimation network design.

\begin{figure*}
\centering
\includegraphics[width=\linewidth]{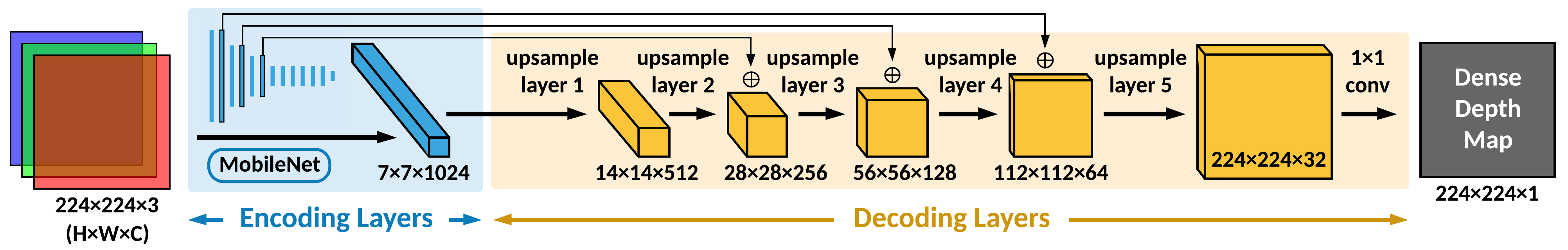}
\caption{Proposed network architecture. Dimensions of intermediate feature maps are given as height ${\times}$ width ${\times}$ \# channels. Arrows from encoding layers to decoding layers denote additive (rather than concatenative) skip connections.}
\label{fig:architecture}
\end{figure*}

\vspace{-10pt}

\section{Methodology}
\label{sec:methodology}

In this section, we describe our proposed network architecture, the motivation behind our design choices, and the steps we take to reduce inference runtime.

\subsection{Network Architecture}

Our proposed fully convolutional encoder-decoder architecture  is shown in~\prettyref{fig:architecture}. The encoder extracts high-level low-resolution features from the input image. These features are then fed into the decoder, where they are gradually upsampled, refined, and merged to form the final high-resolution output depth map. In developing a depth estimation network that can run in real-time, we seek low-latency designs for both the encoder and the decoder.

\subsubsection{Encoder Network}

The encoder used in depth estimation networks is commonly a network designed for image classification. Popular choices include \vgg{-16}~\cite{simonyan2014very} and \resnet{-50}~\cite{he2016deep} because of their strong expressive power and high accuracy. However, such networks also suffer from high complexity and latency, making them unsuitable for applications running in real-time on embedded systems.

Targeting low latency, we employ a state-of-the-art efficient network, \mobilenet~\cite{howard2017mobilenets}, as our encoder of choice. \mobilenet makes use of depthwise decomposition, which factorizes an $m{\times}m{\times}n$ standard convolutional layer into $n$ $m{\times}m$ depthwise layers and a $1{\times}1$ pointwise layer. Since each filter in a depthwise layer only convolves with a \emph{single} input channel, the complexity of a depthwise layer is much lower than that of a standard convolutional layer, where each filter convolves with \emph{all} input channels. Moreover, each pointwise filter is just a $1{\times}1$ kernel, so the number of MACs performed by a pointwise layer is $m^2$ times smaller than that of the original standard convolution. Therefore, depthwise decomposition significantly reduces the complexity of a convolutional layer, making \mobilenet more efficient than networks with standard convolution like \resnet{} and \vgg. This translates to reduced latency, 

\subsubsection{Decoder Network}

The objective of the decoder is to merge and upsample the output of the encoder to form a dense prediction. A key design aspect of the decoder is the upsample operation used (e.g., unpooling, transpose convolution, interpolation combined with convolution).

Our decoder network (termed \nnconv{5}) consists of five cascading upsample layers and a single pointwise layer at the end. Each upsample layer performs $5\times5$ convolution and reduces the number of output channels by $1/2$ relative to the number of input channels. Convolution is followed by nearest-neighbor interpolation that doubles the spatial resolution of intermediate feature maps. Interpolating \emph{after} convolution instead of before lowers the resolution of feature maps processed by the convolutional layers. We use depthwise decomposition to further lower the complexity of all convolutional layers, resulting in a slim and fast decoder.

\subsubsection{Skip Connections}

Encoder networks typically contain many layers to gradually reduce the spatial resolution and extract higher-level features from the input. The output of the encoder into the decoder becomes a set of low resolution features in which many image details can be lost, making it more difficult for the decoder to recover pixel-wise (dense) data. Skip connections allow image details from high resolution feature maps in the encoder to be merged into features within the decoder; this helps the decoding layers reconstruct a more detailed dense output. Skip connections have been previously been used in networks for image segmentation such as U-Net~\cite{ronneberger2015u} and DeeperLab~\cite{yang2019deeperlab}, showing that they can be beneficial in networks producing dense outputs.

We incorporate skip connections from the \mobilenet encoder to the outputs of the middle three layers in the decoder. Feature maps are combined with via addition rather than concatenation, to avoid increasing the number of feature map channels processed by the decoding layers.

\subsection{Network Compilation}

Our proposed network architecture is fully convolutional and makes use of depthwise decomposition in both the encoder and the decoder. Depthwise separable convolutional layers are currently not yet fully optimized for fast runtime in commonly-used deep learning frameworks. This motivates the need for hardware-specific compilation to translate the complexity reduction achievable with depthwise layers into runtime reduction on hardware. We use the TVM compiler stack~\cite{chen2018tvm} to compile our proposed network design for deployment on embedded platforms such as the Jetson TX2.

\subsection{Network Pruning}

To reduce network latency even further, we perform post-training network pruning using the state-of-the-art algorithm, NetAdapt~\cite{yang2018netadapt}. Starting from a trained network, NetAdapt automatically and iteratively identifies and removes redundant channels from the feature maps to reduce the computational complexity.  In each iteration, NetAdapt generates a set of network proposals simplified from a reference network. The network proposal with the best accuracy-complexity trade-off is then chosen and used as the reference network in the next iteration. The process continues until the target accuracy or complexity is achieved. Network complexity can be gauged by indirect metrics (e.g., MACs) or direct metrics (e.g., latency on a target hardware platform).

\section{Experiments} 
\label{sec:experiments}

In this section, we present experiment results to demonstrate our approach. We first present an evaluation against existing work and then provide an ablation study of our design. We offer comparisons of various encoder and decoder options, analysing them based on accuracy and latency metrics. We also show that hardware-specific compilation reduces the runtime cost of the depthwise separable layers within our network and that network pruning helps improve the efficiency of both of the encoder and the decoder.

\subsection{Experiment Setup}
\label{sec:exp-setup}

We train our networks and evaluate their accuracy on the NYU Depth v2 dataset~\cite{silberman2012indoor} with the official train/test data split. Network training is similar to \cite{ma2017sparse} and is implemented in PyTorch~\cite{paszke2017automatic} with 32-bit floating point precision. For training, a batch size of 8 and a learning rate of $0.01$ are used. The optimizer is SGD with a momentum of 0.9 and a weight decay of 0.0001. Encoder weights (e.g., for \mobilenet and \resnet) are pretrained on ImageNet~\cite{deng2009imagenet}. Accuracy is measured by both $\delta_1$ (the percentage of predicted pixels where the relative error is within $25\%$) and \rmse (root mean squared error). For evaluation, a batch size of 1 is used. Precision is kept at 32-bit floating point. Our primary target platform is a Jetson TX2 in max-N mode.\footnote{Max-N mode: all CPU cores in use and GPU clocked at 1.3 GHz.}

\subsection{Final Results and Comparison With Prior Work}

\begin{figure}[h]
\centering
\includegraphics[width=\linewidth]{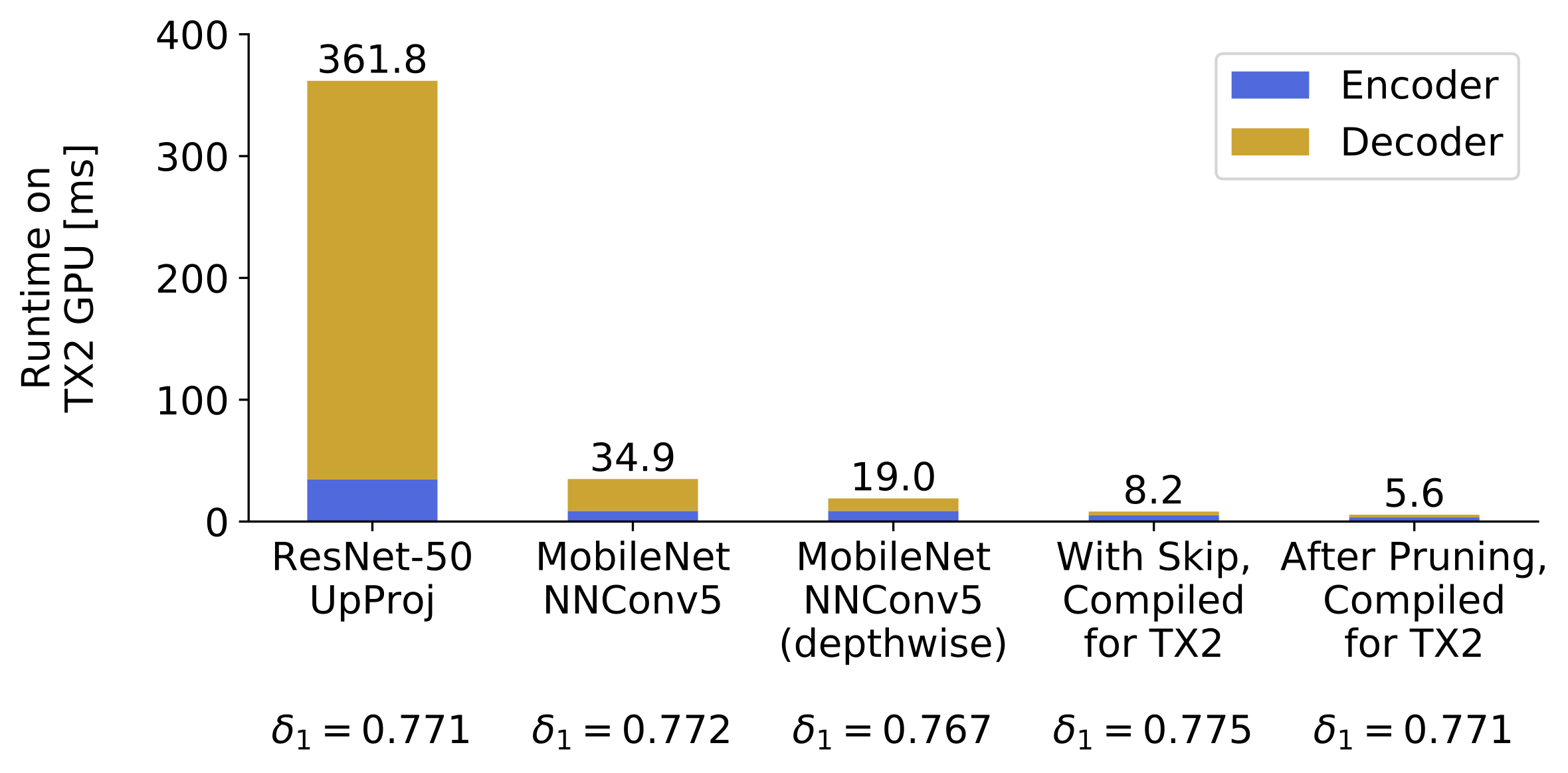}
\caption{Reduction in inference runtime achieved with our approach. Stacked bars represent encoder-decoder breakdown; total runtimes are listed above the bars. The row of $\delta_1$ accuracies listed at the bottom shows the impact of individual steps in our approach on accuracy. Relative to \resnet{-50} with \upproj, our final model achieves 65 times speedup while maintaining accuracy.}
\label{fig:progression}
\end{figure}

Results achieved with our methodology are summarized in~\prettyref{fig:progression}. \resnet{-50} with \upproj serves as a baseline; this network follows the architecture described in~\cite{laina2016deeper}.\footnote{With modifications, namely a $224\times224$ input to the encoder and five (instead of four) upsample layers in the decoder. This is done to match our architecture and allow for a more fair comparison to our network. Consequently, the accuracy and runtime reported in~\prettyref{fig:progression} will differ from that of the unmodified model reported as part of our evaluation in~\prettyref{tab:evaluation}.} The runtime of this baseline network is largely dominated by the decoder. In our approach, the most immediate and significant runtime reduction comes from using a smaller and computationally simpler decoder. However, the decoder continues to dominate network runtime when combined with the \mobilenet encoder. By simplifying the convolutions within the decoder with depthwise decomposition, the runtime of the decoder begins to more closely match that of \mobilenet. Pruning and compiling the network for the target Jetson TX2 platform lower the runtime of both the encoder and the decoder even further, ultimately reducing total inference runtime by a factor of 65 times relative to the baseline and enabling an extremely high throughput of up to 178 fps.

\begin{table}[ht]
\caption{Comparison with prior work. For $\delta_1$, higher is better. For all others, lower is better. Runtimes are measured on a Jetson TX2. Our network design achieves close to state-of-the-art accuracy and is an order of magnitude faster.}
\centering
\scriptsize
\setlength\tabcolsep{3pt} 
\label{tab:evaluation}
\begin{threeparttable}[t]
\begin{tabular}{| l || *{6}{ c |}  }
\hline
\textbf{on NYU Depth v2} & \thead{{Input}\\{Size}} & \thead{{MACs}\\{[G]}} & \rmse & $\delta_1$ & \thead{{CPU}\\{[ms]}} & \thead{{GPU}\\{[ms]}} \\
\hline \hline
Eigen \etal~\cite{eigen2014depth} & 228$\times$304 & 2.06 & 0.907 & 0.611 & 307 & 23 \\
Eigen \etal~\cite{eigen2015predicting} (AlexNet) & 228$\times$304 & 8.39 & 0.753 & 0.697 & 1391 & 96\\ 
Eigen \etal~\cite{eigen2015predicting} (VGG) & 228$\times$304 & 23.4 & 0.641 & 0.769 & 2797 & 195 \\
Laina \etal~\cite{laina2016deeper} (UpConv) & 228$\times$304 & 22.9 & 0.604 & 0.789 & 2384 & 237 \\ 
Laina \etal~\cite{laina2016deeper}  (UpProj) & 228$\times$304 & 42.7 & {\bf0.573} & {\bf0.811} & 3928 & 319 \\
Xian \etal~\cite{xian2018monocular} & 384$\times$384 & 61.8 & 0.660 & 0.781 & 4429 & 283 \\
\hline \hline
This Work & 224$\times$224 & {\bf0.37} & 0.604 & 0.771 & {\bf37} & {\bf5.6} \\
\hline
\end{tabular}
\end{threeparttable}
\end{table}

Accuracy and latency metrics of our model in comparison with prior work\footnote{MACs and runtimes were generated from our re-implemented models.} are summarized in \prettyref{tab:evaluation}. We evaluate against depth estimation methods that use deep learning but do not involve additional processing such as CRFs, since the additional processing incurs computation and runtime cost. We measure the active power consumption when running our model on the TX2 in max-N mode to be under 10 W. We additionally report runtime and power consumption data for the TX2 in the more energy-efficient max-Q mode.\footnote{Max-Q mode: only the ARM Cortex-A57 cores in use and GPU clocked at 0.85 GHz. Configured for best power-throughput tradeoff.} ~\prettyref{tab:platforms} summarizes this data. We note that with the TX2 in max-Q mode, our model can achieve close to real-time speeds on the CPU and can still easily surpass real-time speeds on the GPU, with active power consumption under 5 W.

\begin{table}[ht]
\caption{Inference runtime and peak power consumption when deploying our model on the Jetson TX2 in high performance (max-N) and high energy-efficiency (max-Q) modes. Active power consumption can be estimated by subtracting the reported idle power consumption.}
\centering
\scriptsize
\setlength\tabcolsep{6pt} 
\label{tab:platforms}
\begin{threeparttable}[t]
\begin{tabular}{| l || c | c | r | }
\hline
Platform & Runtime & Max Frame Rate & Power Consumption \\ 
\hline \hline
TX2 GPU (max-N) & 5.6 ms & 178 fps & 12.2 W (3.4 W idle)\\
TX2 GPU (max-Q) & 8.2 ms & 120 fps & 6.5 W (1.9 W idle) \\
\hline
TX2 CPU (max-N) & 37 ms & 27 fps & 10.5 W (3.4 W idle) \\
TX2 CPU (max-Q) & 64 ms & 15 fps & 3.8 W (1.9 W idle) \\
\hline
\end{tabular}
\begin{tablenotes}\footnotesize
\end{tablenotes}
\end{threeparttable}
\end{table}

\prettyref{fig:examples} visualizes results of depth estimation produced by our model on images from the NYU Depth v2 dataset. Skip connections between encoding and decoding layers improve the sharpness and visual clarity of depth map outputs, while network pruning preserves and even enhances clarity. We also show an error map visualizing the difference between the output of our model and ground truth, noting that the error is highest at boundaries and at distant objects.

\begin{table}
\begin{center}
\setlength\tabcolsep{2pt} 
\begin{tabular}{ cccccc }

\includegraphics[width=0.15\linewidth]{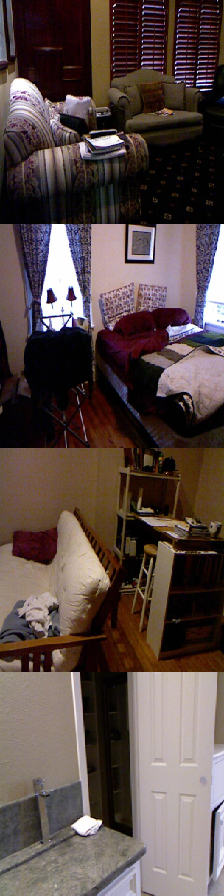} &
\includegraphics[width=0.15\linewidth]{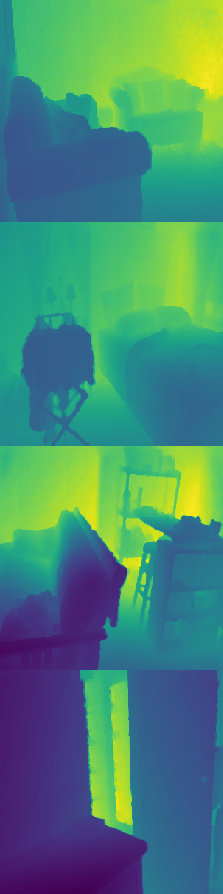} &
\includegraphics[width=0.15\linewidth]{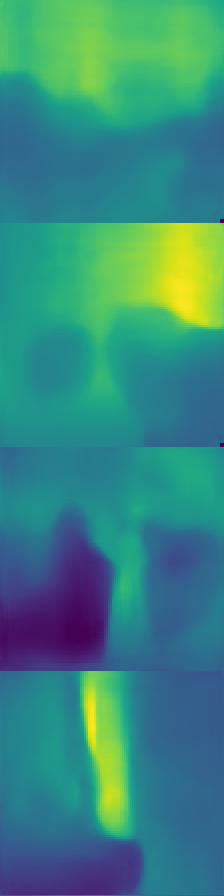} &
\includegraphics[width=0.15\linewidth]{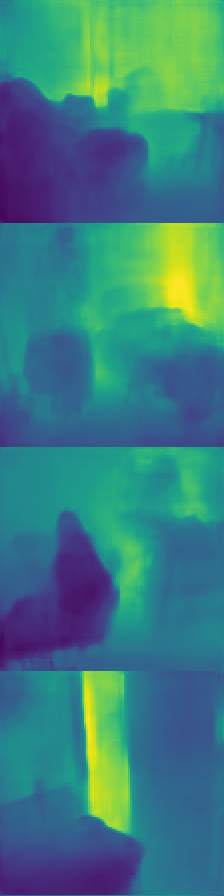} &
\includegraphics[width=0.15\linewidth]{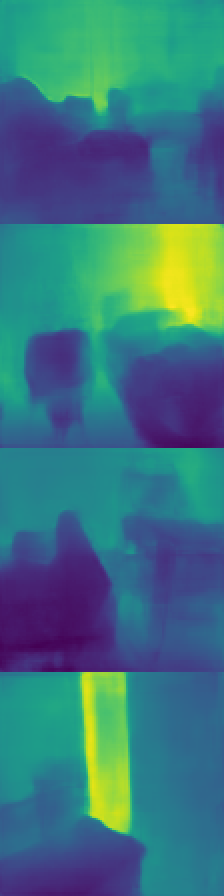} &
\includegraphics[width=0.15\linewidth]{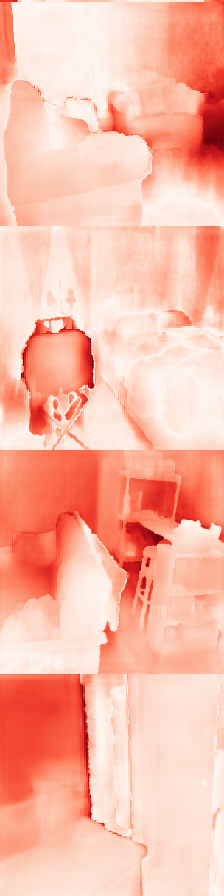} \\
(a) & (b) & (c) & (d) & (e) & (f)

\end{tabular}
\end{center}
\captionof{figure}{Visualized results of depth estimation on the NYU Depth v2 dataset. (a) input RGB image; (b) ground truth; (c) our model, without skip connections, unpruned; (d) our model, with skip connections, unpruned; (e) our model, with skip connections, pruned; (f) error map between the output of our final pruned model and ground truth, where redder regions indicate higher error.} \label{fig:examples}
\end{table}

\subsection{Ablation Study: Encoder Design Space}
\label{sec:encoder_design_space}

A common encoder used in existing high-accuracy approaches~\cite{laina2016deeper, xian2018monocular} is \resnet{-50}~\cite{he2016deep}. Targeting lower encoder latency, we consider the smaller \resnet{-18}~\cite{he2016deep} and \mobilenet~\cite{howard2017mobilenets} as alternatives to \resnet{-50}. The last average pooling layer and fully connected layers are removed from the \mobilenet{} and \resnet{} architectures. To make the encoders compatible with a fixed decoder structure, we append a $1{\times}1$ convolutional layer to the end of \resnet{} encoders, such that the output from all encoder variants has a consistent shape of $7{\times}7$ with $1024$ channels.

We compare all three encoder options against each other in \prettyref{tab:encoder}. The reported runtimes are obtained by compiling and running the encoder networks in PyTorch. Runtimes for \resnet{-50} and \resnet{-18} are too high, even on the TX2 GPU, to achieve real-time speeds (i.e., above 30 fps) if these encoders are paired with decoders of similar latency. In comparison, \mobilenet efficiently trades off between accuracy and latency, and has a noticeably lower GPU runtime. We therefore select \mobilenet as our encoder.

We note that despite its lower complexity, \mobilenet is an order of magnitude slower on the TX2 CPU than \resnet{-18}. This can be attributed to as-of-yet unoptimized implementations for depthwise layers in deep learning frameworks, motivating the need for an alternate deep learning compiler, as will be discussed in \prettyref{sec:exp-compilation}. 

\begin{table}[ht]
\caption{Comparison of encoders. \rmse and $\delta_1$ are for encoder-decoder networks with the decoder fixed as \nnconv{5}. All other metrics are for the encoder in isolation. Runtimes are measured on a TX2. \mobilenet is selected as best encoder option.}
\centering
\scriptsize
\setlength\tabcolsep{5pt} 
\label{tab:encoder}
\begin{threeparttable}[t]
\newcolumntype{M}{>{$\vcenter\bgroup\hbox\bgroup}c<{\egroup\egroup$}}
\begin{tabular}{| l || *{6}{ c |}  }
\hline
\textbf{Encoder} & \thead{{Weights}\\{[M]}} & \thead{{MACs}\\{[G]}} & \thead{{\rmse}\\{[meters]}} & $\delta_1$ & \thead{{CPU}\\{[ms]}} & \thead{{GPU}\\{[ms]}} \\
\hline \hline
\resnet{-50} & 25.6 & 4.19 & {\bf0.568} & {\bf0.800} & 610 & 35.0 \\
\resnet{-18} & 11.7 & 1.84 & {\bf0.568} & 0.782 & {\bf220} & 15.2 \\
\mobilenet & {\bf3.19} & {\bf0.57} & 0.579 &  0.772 & 3700 & {\bf8.7} \\
\hline
\end{tabular}
\end{threeparttable}
\end{table}

\subsection{Ablation Study: Decoder Design Space}
\label{sec:decoder_design_space}

While encoders have been well characterized in deep learning applications, decoders have been less extensively explored, especially in the context of efficient network design. We consider several decoder design aspects: upsample operation, depthwise decomposition, and skip connections.

\subsubsection{Upsample Operation}
\begin{figure}[t]
\includegraphics[width=\linewidth]{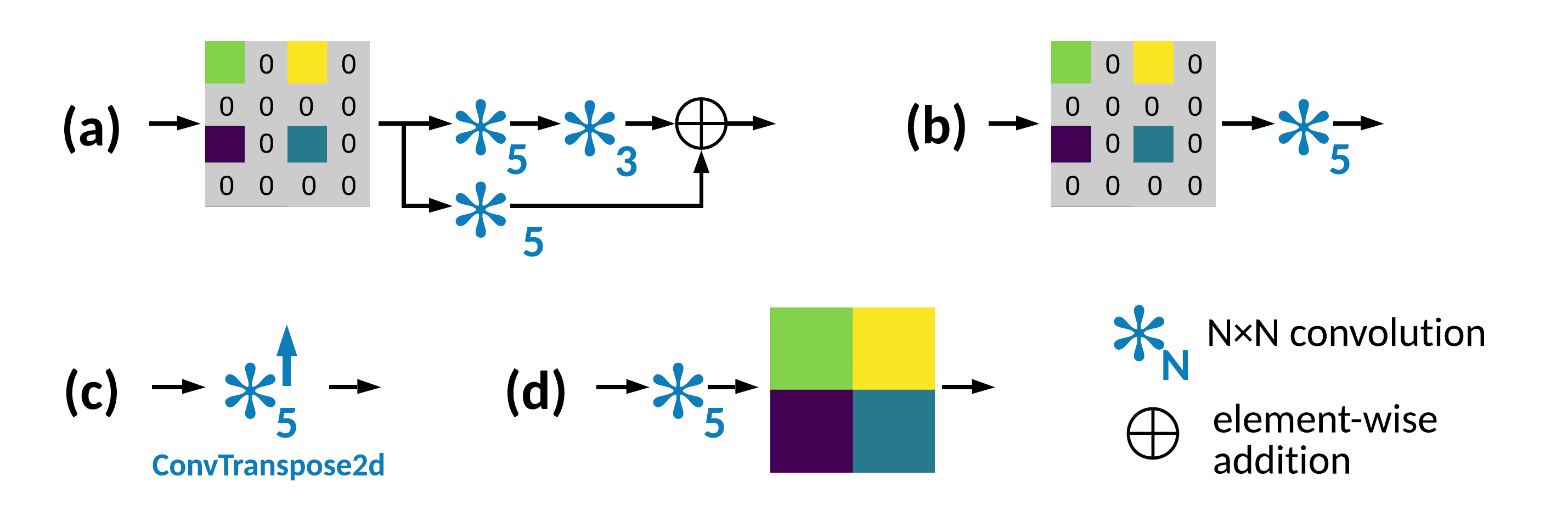}
\caption{Visual representations of different upsample operations.\\(a) \upproj, (b) \upconv, (c) \deconv{5}, (d) \nnconv{5}.}
\label{fig:upsample}
\end{figure}

We survey four ways of upsampling in the decoder. Their characteristics are listed below, and visual representations are shown in~\prettyref{fig:upsample}:
\begin{enumerate}[label=(\alph*),start=1]
    \item \upproj~\cite{laina2016deeper}: $2\times2$ unpooling (zero-insertion) followed by a two-branched residual structure that computes a total of three convolutions (two $5\times5$ and one $3\times3$).
    \item \upconv~\cite{laina2016deeper}: $2\times2$ unpooling (zero-insertion) followed by a single 5$\times$5 convolution.
    \item \deconv{5}: transpose convolution using a $5\times5$ kernel.\footnote{We use a kernel size of 5 to fairly compare against \upconv.}
    \item \nnconv{5}: $5\times5$ convolution followed by nearest-neighbor interpolation\footnote{An alternate option would be bilinear interpolation. We select nearest-neighbor interpolation as is it a simpler operation with more consistent implementations across deep learning frameworks and compilers.} with a scale factor of 2.
\end{enumerate}

We implement four decoder variants using these upsample operations, keeping the structure fixed at 5 decoding layers with $1\times1$ convolution at the end.
\prettyref{tab:upsample} compares the four decoders. \upproj is most complex, due to its larger number of convolutions per upsample layer. It achieves the highest $\delta_1$ accuracy but is the slowest. \upconv is less complex and faster than \upproj, but its CPU and GPU runtimes are still too slow for real-time processing. \deconv{5} has an identical number of weights and MACs as \upconv and is noticeably faster on both the CPU and GPU. However, it can be prone to introducing checkerboard artifacts in its outputs~\cite{odena2016deconvolution}, which helps explain its lower accuracy. \nnconv{5} achieves higher $\delta_1$ accuracy and lower \rmse than both \upconv and \deconv{5}, with a slightly lower GPU runtime. We therefore select \nnconv{5} as our decoder.

\begin{table}[ht]
\caption{Comparison of decoders. \rmse and $\delta_1$ are for encoder-decoder networks with a \mobilenet encoder. All other metrics are for the decoder in isolation. Runtimes are measured on a TX2. \nnconv{5} is selected as best decoder option.}
\centering
\scriptsize
\setlength\tabcolsep{4pt} 
\label{tab:upsample}
\begin{threeparttable}[t]
\begin{tabular}{| l || *{6}{ c |}  }
\hline
\textbf{Decoder} & \thead{{Weights}\\{[M]}} & \thead{{MACs}\\{[G]}} & \thead{{\rmse}\\{[meters]}} & $\delta_1$ & \thead{{CPU}\\{[ms]}} & \thead{{GPU}\\{[ms]}} \\
\hline \hline
(a) \upproj~\cite{laina2016deeper} & 38.1 & 28.0 & 0.599 & {\bf0.774} & 3300 & 325 \\
(b) \upconv~\cite{laina2016deeper} & {\bf17.5} & 12.9 & 0.591 & 0.771 & 1600 & 238 \\ \hline
(c) \deconv{5} & {\bf17.5} & 12.9 & 0.596 & 0.766 & {\bf290} & 31.0 \\
(d) \nnconv{5} & {\bf17.5} & {\bf3.21} & {\bf0.579} & 0.772 & 410 & {\bf26.2} \\
\hline
\end{tabular}
\end{threeparttable}
\end{table}

\subsubsection{Depthwise Separable Convolution}

After selecting \mobilenet as our encoder and \nnconv{5} as our decoder, we observe that the runtime of our network is dominated by the decoder (see~\prettyref{fig:progression} for the encoder-decoder breakdown). This motivates us to simplify our decoder even further. Similar to how depthwise decomposition lowers the complexity and latency in \mobilenet, we now replace all convolutions within the decoder with depthwise separable convolutions. 

\prettyref{tab:skip} shows that depthwise decomposition in the decoder lowers inference runtime on the GPU by almost half.\footnote{In contrast, runtime on the CPU increases, despite reduced number of MACs. This is due to the inefficient CPU operation of depthwise layers.} However, as 
was the case with \mobilenet, depthwise layers in the decoder result in a slight accuracy loss, due to the reduction in trainable parameters and computation. In order to restore some of the lost accuracy, we incorporate skip connections between the encoding and decoding layers.

\begin{table}[ht]
\caption{Impact of depthwise decoding layers and skip connections on network complexity and TX2 runtime.}
\centering
\scriptsize
\setlength\tabcolsep{3pt} 
\label{tab:skip}
\begin{threeparttable}[t]
\newcolumntype{M}{>{$\vcenter\bgroup\hbox\bgroup}c<{\egroup\egroup$}}
\begin{tabular}{| l || *{6}{ M |}  }

\hline
\mobilenet-\nnconv{5} & \thead{{Weights}\\{[M]}} & \thead{{MACs}\\{[G]}} & \thead{{\rmse}\\{[meters]}} & $\delta_1$ & \thead{{CPU}\\{[ms]}} & \thead{{GPU}\\{[ms]}} \\
\hline \hline
with standard decoder & 20.6 & 3.78 & {\bf0.579} & 0.772 & {\bf4100} & 34.9 \\
\hline
with depthwise decoder & {\bf3.93} & {\bf0.74} & 0.584 & 0.767 & 5200 & {\bf18.6} \\
\hline \hline
depthwise \& skip-concat & 3.99 & 0.85 & 0.601 & {\bf0.776} & 5500 & 26.8 \\
\hline
depthwise \& skip-add & {\bf3.93} & {\bf0.74} & 0.599 & 0.775 & 5100 & 19.1 \\
\hline

\end{tabular}
\end{threeparttable}
\end{table}

\subsubsection{Skip Connections}

We consider both additive and concatenative skip connections. 
Concatenative skip connections increase the computational complexity of the decoder since decoding layers need to process feature maps with more channels. ~\prettyref{tab:skip} shows that this improves the $\delta_1$ accuracy but also noticeably increases CPU and GPU runtimes. In contrast, using additive skip connections leaves the number of channels in the decoder unchanged and has a negligible impact on inference runtime while achieving almost the same accuracy boost. We therefore use additive skip connections in our final network design. As shown in \prettyref{fig:examples}(d), skip connections noticeably improve the sharpness and visual clarity of the depth maps output by our network design.

\subsection{Hardware-Specific Optimization}
\label{sec:exp-compilation}

Current deep learning frameworks rely on framework-specific operator libraries, where the level of hardware-specific optimization of operator implementations may vary. Our proposed network architecture incorporates depthwise layers throughout the encoder and decoder. These layers are currently not yet fully optimized in commonly-used deep learning frameworks. As a result, although depthwise decomposition significantly reduces the number of \mac\ in a network, a similar reduction is not reflected in latency. The left portion of Table~\ref{tab:tvm} highlights exactly this: the TX2 CPU runtime of \mobilenet-\nnconv{5} is high to begin with, due to the prevalence of depthwise layers in \mobilenet, and it increases even more when we use depthwise layers in the decoder. To address the observed runtime inefficiencies of depthwise layers, we use the TVM compiler stack~\cite{chen2018tvm}. TVM performs hardware-specific scheduling and operator tuning that allows the impact of reduced operations to be translated into reduced processing time. The right portion of Table~\ref{tab:tvm} reports TX2 runtimes for networks compiled with TVM. Depthwise decomposition in the decoder now reduces CPU runtime by 3.5 times and GPU runtime by 2.5 times. 

\begin{table}[t]
\centering
\scriptsize
\setlength\tabcolsep{4pt} 
\caption{Hardware-specific compilation enables fast depthwise layers in our network. Runtimes are measured on the TX2.}
\label{tab:tvm}
\begin{threeparttable}[t]
\newcolumntype{M}{>{$\vcenter\bgroup\hbox\bgroup}c<{\egroup\egroup$}}

\begin{tabular}{| l || M | M || M | M | }
\hline
\multirow{3}{*}{\mobilenet-\nnconv{5}} & \multicolumn{2}{M ||}{in PyTorch} & \multicolumn{2}{M |}{using TVM} \\ \cline{2-3} \cline{4-5} 
& CPU [ms] & GPU [ms] & CPU [ms] & GPU [ms]
\\ \hline \hline
with standard decoder & {\bf4100} & 34.9 & 176 & 20.9 \\
with depthwise decoder & 5200 & {\bf18.6} & {\bf50} & 8.3 \\
\hline
with depthwise \& skip-add & 5100 & 19.1 & 66 & {\bf8.2} \\
\hline

\end{tabular}
\end{threeparttable}
\end{table}

\subsection{Network Pruning}
\label{sec:pruning}

Prior to network pruning, our architecute (\mobilenet-\nnconv{5} with depthwise decomposition in the decoder and additive skip connections) already surpasses real-time throughput on the TX2 GPU but does not yet achieve real-time speeds on the TX2 CPU. Network pruning lowers the model's runtime so that it can achieve a CPU framerate above 25 fps that is more suitable for real-time inference. As shown in ~\prettyref{tab:pruning}, pruning achieves a 2 times reduction in \mac, a 1.5 times reduction in GPU runtime, and a 1.8 times reduction in GPU runtime with almost the same accuracy. ~\prettyref{fig:examples}(e) shows that pruning process preserves the sharpness and visual clarity of output depth maps.

\begin{figure}[htbp]
\centering
\includegraphics[width=\linewidth]{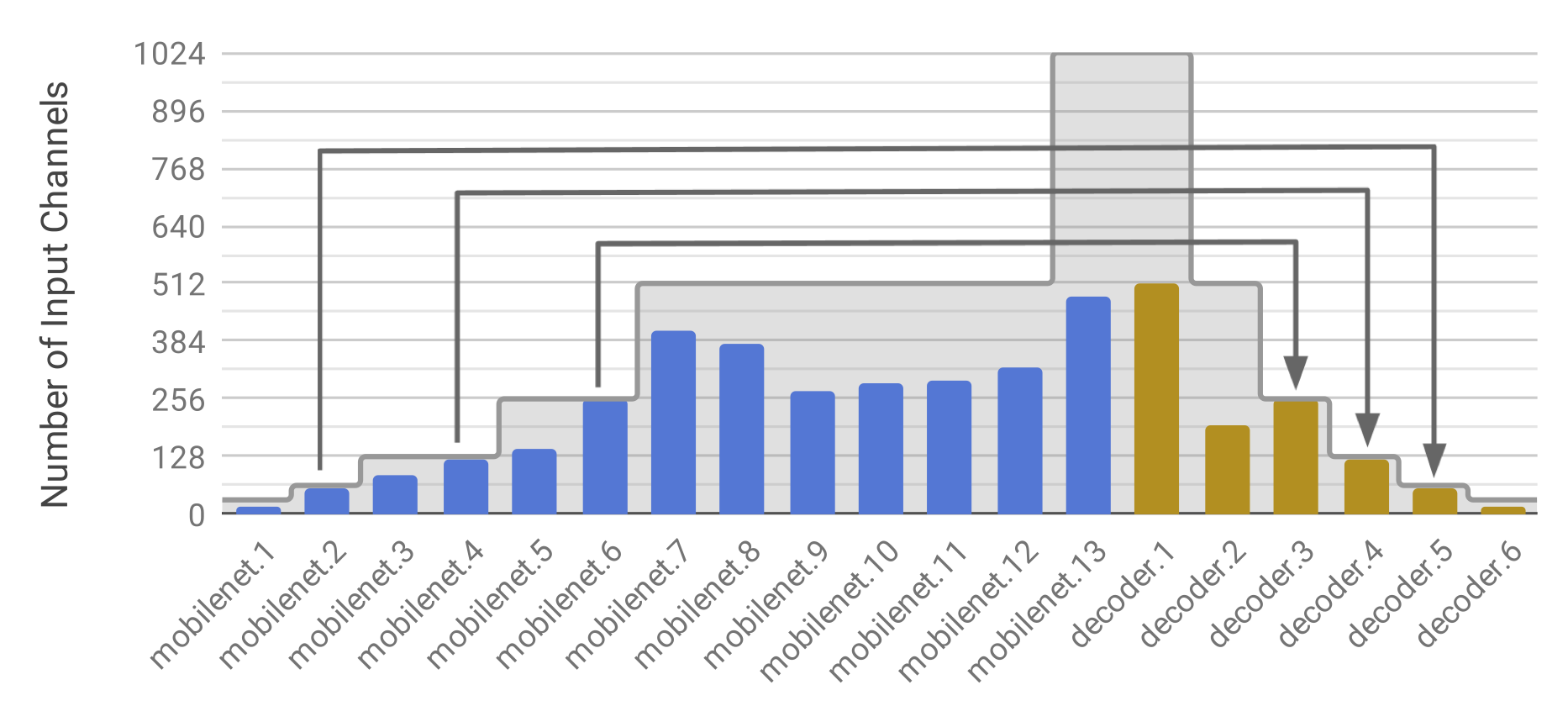}
\caption{Number of input channels to each layer in our network architecture after pruning. The shaded part represents the architecture before pruning. The very first layer to the network (\texttt{mobilenet.0}) is not shown since the channel size of the input fed into the network remains fixed at 3 channels (RGB).}
\label{fig:pruning_chns}
\end{figure}

Fig.~\ref{fig:pruning_chns} shows the pruned architecture. We can see that there are two bottlenecks: one in the encoder (the layer \texttt{mobilenet.9}) and one in the decoder (the layer \texttt{decoder.2}). This is consistent with the observations in~\cite{yang2018netadapt,sandler2018mobilenetv2}.

\begin{table}[ht]
\centering
\scriptsize
\caption{Impact of pruning on our encoder-decoder network. Runtimes are measured post-compilation for the TX2.}
\label{tab:pruning}
\begin{threeparttable}[t]
\newcolumntype{M}{>{$\vcenter\bgroup\hbox\bgroup}c<{\egroup\egroup$}}
\begin{tabular}{| M || *{3}{ M |}  }
\hline
& Before Pruning & After Pruning & Reduction \\ 
\hline \hline
Weights & 3.93M & 1.34M & 2.9$\times$ \\
\mac & 0.74G & 0.37G & 2.0$\times$ \\
\hline
\rmse & 0.599 & 0.604 & - \\
$\delta_1$ & 0.775 & 0.771 & - \\
\hline
CPU [ms] & 66 & 37 & 1.8$\times$ \\
GPU [ms] & 8.2 & 5.6 & 1.5$\times$ \\
\hline
\end{tabular}
\end{threeparttable}
\end{table}

\vspace{-10pt}

\section{Conclusion} 
\label{sec:conclusion}

In this work, we enable high-speed depth estimation on embedded systems. We achieve high frame rates by developing an efficient network architecture, with a low-complexity and low-latency decoder design that does not dominate inference runtime even when combined with a small \mobilenet encoder. The size of our compact model is further reduced by applying a state-of-the-art pruning algorithm. Hardware-specific compilation is used to translate complexity reduction into lower runtime on a target platform. On the Jetson TX2, our final model achieves runtimes that are an order of magnitude faster than prior work, while maintaining comparable accuracy.

Although this work focuses on depth estimation, we believe that similar approaches can be used to achieve real-time performance with deep-learning based methods for other dense prediction tasks, such as image segmentation.




\clearpage
\bibliographystyle{IEEEtranN}

\end{document}